\documentclass{article}

\usepackage{arxiv}

\usepackage[utf8]{inputenc} 
\usepackage[T1]{fontenc}    
\usepackage{hyperref}       
\usepackage{url}            
\usepackage{booktabs}       
\usepackage{amsfonts}       
\usepackage{nicefrac}       
\usepackage{microtype}      
\usepackage{cleveref}       
\usepackage{lipsum}         
\usepackage{graphicx}
\usepackage{natbib}
\usepackage{doi}
\usepackage{relsize}

\title{The most general manner  to injectively  align  true and predicted  segments}

\author{ \href{https://orcid.org/0000-0003-3255-3729}{\includegraphics[scale=0.06]{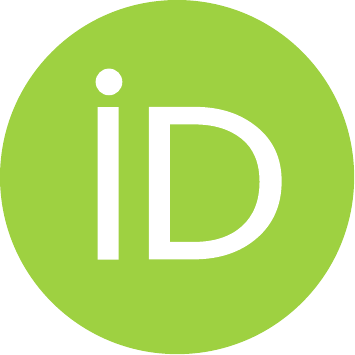}\hspace{1mm}Maarten Marx}\thanks{This research was supported in part by the
 Netherlands Organization for Scientific Research (NWO)  through the ACCESS project grant  CISC.CC.016, and by the University of Amsterdam through Humane AI.} \\
	Informatics Institute\\
	 University of Amsterdam\\
	The Netherlands \\
	\texttt{maartenmarx@uva.nl} 
}

\renewcommand{\shorttitle}{ {Aligning true and predicted segments}}

\begin{document}
\maketitle

\begin{abstract}

Kirilov et al (2019) develop a metric, called \emph{Panoptic
Quality (PQ)}, to evaluate image segmentation methods. The metric is
based on a confusion table, and compares a predicted   to a
ground truth segmentation.  
The only non straightforward part in this comparison 
is to
 align the  segments in the two segmentations.  
 A metric only works well if that alignment   is a
{partial bijection}.    Kirilov et al (2019)  list 3 desirable properties for a definition of  alignment: it should be simple, interpretable and effectively computable.  There are many definitions guaranteeing a
partial bijection and these 3 properties.  We present the weakest:
one that is both sufficient and necessary to guarantee that the alignment is a  partial bijection. 
This new condition is  effectively computable and
natural. It simply says that the number of correctly predicted
elements (in image segmentation, the pixels) should be larger than the number of missed,
and larger than the number of spurious elements.
This is strictly weaker
than the proposal in   Kirilov et al (2019). 
In formulas, instead of  \textbar TP\textbar{} \textgreater{}
\textbar FN\textbar{} +\textbar FP\textbar,  the weaker condition  requires that 
\textbar TP\textbar{} \textgreater{} \textbar FN\textbar{} and
\textbar TP\textbar{} \textgreater{} \textbar FP\textbar{}.
We evaluate the new alignment condition  theoretically and empirically. 

\end{abstract}


\hypertarget{introduction}{%
\section{Introduction}\label{introduction}}

 \cite{kiri:pano19} develop a metric, called \emph{Panoptic Quality
(PQ)}, to evaluate image segmentation methods. 
PQ is developed for images, but it can be used for text as
well, and even for any clustering problem. The only requirement is that
there is an underlying set of elements (in images the pixels, in text
segmentation typically tokens) which are \emph{partially partioned}
(i.e., elements are combined into non overlapping segments, but not all
elements need to be assigned to a segment).

PQ is especially developed for segmentation problems in which
\emph{exact matches} are unfeasible and not even needed for
successfull applications. PQ is a very elegant metric which can be seen
as combining the \emph{segmentation quality} (how well do predicted and
true segments ``match/overlap''?) with the \emph{recognition quality}
(the usual precision and recall questions).

The only non straightforward part  in comparing a predicted to a true segmentation is to
match   the separate segments in the two segmentations. Apart
from the fact that they are both subsets of the same underlying domain
there is no other information, and this matching can thus be done in
many different ways.

Kirilov et al (2019) require a simple, efficient and interpretable
matching process. Besides that, the alignment should be   a \textbf{partial bijection}: for every predicted segment
$h$ there is at most one true segment $t$ and vice-verse. In Theorem
1 they show that the natural condition stating that $h$ and $t$ match
iff $IoU(h,t)>.5$ guarantees a partial bijection.

Having this it is straightforward to define PQ. And indeed the thus
defined metric is effectively computable, interpretable and simple, as required.
Great! Except\ldots.., there are more, even many more, definitions of
matching which guarantee a partial bijection, and are also interpretable
and effectively computable. Why would we chose $IoU(h,t)>.5$?

\hypertarget{when-to-match}{%
\section{When to match?}\label{when-to-match}}

The condition $IoU(h,t)>.5$ is equivalent to
\[ \vert h\cap t \vert > \vert h\oplus t \vert,\] which says that the
overlap between $h$ and $t$ is larger than their symmetric
difference.
The symmetric difference consists of the two (disjoint)
\emph{error regions}, the \emph{missed pixels} and the \emph{spurious
pixels}. (Or the false negative and false positive pixels). Stated like
this, the condition is indeed both interpretable and reasonable. We can
paraphrase it as ``there are more overlapping than erroneously mapped
pixels''. As the two error regions are disjoint, the following condition
is also equivalent:
\[ \vert h\cap t \vert > \vert \mathit{missed}(h,t) \vert + \vert \mathit{spurious}(h,t) \vert. \]
Here $\mathit{missed}(h,t)=t\setminus h$ and
$\mathit{spurious}(h,t) = h\setminus t$ are  the sets of
false negative and false positive pixels, respectively.

Our research question now is

\begin{quote}
Are there other useful\footnote{We added the property \emph{useful}, because   the identity matching satisfies all other properties, but obviously that is often too strict and thus  not (very) useful.}, interpretable, simple and effective matching definitions
which imply the partial bijection property? And if so, is there a most
general one?
\end{quote}

Indeed there is a simpler condition which is sufficient to guarantee a
partial bijection and which is also \textbf{neccessary}: every
``reasonable'' partial bijection satisfies the condition. The condition
is
\[ \vert h\cap t \vert > \vert \mathit{missed}(h,t) \vert  \;\mathbf{\ and\ } \;\vert h\cap t \vert > \vert \mathit{spurious}(h,t) \vert. \;\;\;\;\;\;\;\;\; (*)\]
Below we will develop what ``reasonable'' is. But first we look deeper  into the new condition. Clearly it is simple, natural  and interpretable. 
One may argue that it is conceptually simpler than the  $IoU>.5$ check, because   adding the spurious and missed pixels is no longer needed.

\hypertarget{simple-results}{%
\section{Results about the (*) alignment}\label{simple-results}}

Before we state and prove the characterization theorem we show that alignment defined by  (*) is effective, always a partial bijection and  strictly weaker than the $IoU>.5$ alignment. 

\textbf{Claim 1} Given $H$ and $T$, the condition (*) is effectively
computable in $|H|$ and $|T|$.

\textbf{Proof} A trivial nested for loop over $H$ and $T$ finds the alignment. This can be optimized    using the order on the elements of the underlying domain.

\textbf{Claim 2} Let $H$ and $T$ be two segmentations of the same
set. Then the alignment between segments $h\in H$ and $t\in T$
defined by (*) is a partial bijection.

In proofs it is easier to use this   version of (*):

\[ \vert h\cap t \vert > \vert  t\setminus h  \vert   \;\mathbf{\ and\ } \;\vert h\cap t \vert > \vert  h\setminus t \vert. \;\;\;\;\;\;\;\;\; (*)\]

\textbf{Proof} Suppose not. We will derive a contradiction.
Then there are two possibilities. We treat
one: there is a $t\in T$ and two different (and thus dfisjoint)
$h_1,h_2\in H$ satisfying (*). Thus by assumption, both $|h_1\cap t|>|t\setminus h_1|$
and $|h_2\cap t|>|t\setminus h_2|$.

Because $h_1$ and $h_2$ are disjoint, $h_2\cap t \subseteq t\setminus h_1$ and thus
\[ |t\setminus h_1| \geq |h_2\cap t|.\] 
By assumption
\[|h_2\cap t| >|t\setminus h_2|,\] 
and by the disjointness again
\[|t\setminus h_2| \geq |h_1\cap t|,\] and thus
$|t\setminus h_1|>|h_1\cap t|$, which contradicts with
$|h_1\cap t|>|t\setminus h_1|$. Note that we only used one of the
conjuncts of (*). For the other case, we need the other conjunct.

\textbf{Claim 3} Let $b:H\longrightarrow T$ be the partial bijection
defined by (*). Then for each $h \in dom(b)$,
$IoU(h,b(h))> \frac{1}{3}$.

\textbf{Proof} Denote the three sets in (*) by TP, FP and FN. All three
are disjoint. Now the $IoU$ is equal to $|TP|$ divided by
$|TP|+|FP|+|FN|$. The claim is now immediate.

\textbf{Claim 4} (*) is strictly weaker than the original PQ matching
condition of $IoU(h,t)>.5$.

\textbf{Proof} Because the spurious and missing parts are disjoint it is
weaker. For strictness let $H=\{\{1\}, \{2,3,4\}\}$ and
$T=\{\{1,2,3\}, \{4\}\}$. Then with (*), $\{2,3,4\}$ is aligned to
$\{1,2,3\}$ because it has 2 elements in the overlap, and it has one
missing and one spurious element. But the $IoU$ of these two segments
is equal to $\frac{2}{4}$ and thus not strictly larger than $.5$,
and thus with the original PQ matching condition of $IoU(h,t)>.5$ the
aligment is empty.

Thus the PQ of this pair $H,T$ with the original alignment definition
is $0$, while with alignment according to (*) $TP$ contains the pair
$(\{2,3,4\},\{1,2,3\})$, $FP$ contains the missed segment $\{1\}$
and $FN$ the missed segment $\{4\}$. Thus
\[PQ^* = \frac{IoU(\{2,3,4\},\{1,2,3\})}{|TP|+.5(|FP|+|FN|)}=
\frac{\frac{2}{4}}{1+.5(1+1)}= \frac{1}{4}.\]

Note that for the true segmentation $T=\{\{1,2,3\}, \{4\}\}$, there
are 5 non empty alignments defined by $IoU(h,t)>.5$, and one more, the
one above, defined by (*).

In the next claim we use SQ, RQ and PQ as in \cite{kiri:pano19}. 
Thus SQ is the mean $IoU$ of the True Positives, RQ is the traditional F1 value and PQ is SQ times RQ.

\textbf{Claim 5} There is an extreme case where

\begin{itemize}
\item
  PQ and RQ with the $IoU>.5$  alignment are both 0, as there are no True
  Positives.
\item
  RQ with the (*) alignment approaches 1.
\item
  SQ (and hence also PQ) with the (*) alignment approaches (from above)
  $\frac{1}{3}$.
\end{itemize}

\textbf{Proof} The case is an enlarged variant of the example from Claim 4. 
Let
$c$ and $n$ be two integers, with $c$ even. The true segmentation
consists of a segment of length $\frac{c}{2}$ followed by $n$
segments of length $c+1$. The predicted segmentation is simply the
reverse of the true one. Now all the long segments can be aligned to
each other because for each one there is another with $\frac{c}{2}+1$
elements overlap and thus only $\frac{c}{2}$ missing and
$\frac{c}{2}$ spurious elements. Thus $|TP|=n-2$, $|FP|=1$ and
$|FN|=1$. As $n$ gets large, the $RQ=\frac{n-2}{n-2 + .5(1+1)}$
approaches 1. As $c$ gets large, the IoU of each match approaches $
\frac{.5c +1}{.5c+1 + .5c +.5c}=\frac{1}{3}$ from above, and hence also
SQ.

\hypertarget{the-characterization}{%
\section{The characterization}\label{the-characterization}}

There are bijections between $H$ and $T$ for which (*) does not
hold, but these are in an intuitive sense not \emph{reasonable}. Here is
one over 6 elements:
\begin{center}

\begin{tabular}{l|l|l|}

$H$ & 1,2,3 & 4,5,6 \\
$T$ & 1,2,3,4,5 & 6 \\
\end{tabular}
 \end{center}
This is a total bijection between two segmentations of a linear order.
The two first segments match even with the $IoU$ larger than .5
constraint. The two last segments do not, but are matched just ``because
there is room left''.

We now develop the definition disallowing such matchings. Let $H$ and
$T$ be segmentations of $E$ and $b:H\longrightarrow T$ a partial
bijection. We call such a $H,T,b$ an \emph{alingnment}.
We will use $dom(b)$ and $ran(b)$ to denote the domain
and range of $b$ respectively.

\textbf{Definition 1}
Let $H,T,b$ be an alingnment.
\begin{enumerate}
\def\labelenumi{\arabic{enumi}.}

\item
   We say that $H,T,b$ is \emph{not fair}    if there is an 
   $h\in dom(b)$ and there is a $t\in T$ and  $t\not\in ran(b)$ such that  
  $\vert h\cap t\vert \geq \vert h\cap b(h)\vert$.
\item
  We say that $H,T,b$ is \emph{not fair} if $T,H,b^{-1}$ is not
  fair.
\end{enumerate}

\textbf{Definition 2}
Let $H,T',b'$ be an alingnment.
\begin{enumerate}
\def\labelenumi{\arabic{enumi}.}
\item
  $H,T',b'$ is an \emph{improvement} of the alignment $H,T,b$ if
  $dom(b)=dom(b')$ and for all $h\in dom(b)$ it holds that  
  $h\cap b'(h) \supseteq h\cap b(h)$.
\end{enumerate}


\begin{enumerate}
\def\labelenumi{\arabic{enumi}.}
\setcounter{enumi}{1}
\item
  $H',T,b'$ is an \emph{improvement} of $H,T,b$ if $T,H',b'^{-1}$
  is an improvement of $T,H,b^{-1}$.
\end{enumerate}

Thus with an improvement, we may change one of the segmentations  and either the
range or the domain of the alignment $b$, but only if the $IoU$ remains the same
or increases. 
The following new $T'$ and now partial bijection
$b'$ mapping $\{1,2,3\}$  to itself in $T'$, and $b'(\{4,5,6\})$ remains $\{6\}$ is an improvement
of the example $H,T,b$  given earlier:

\begin{center}

\begin{tabular}{l|l|l|l|}

$H$ & 1,2,3 & &4,5,6 \\
$T$ & 1,2,3,4,5 & &6 \\
$T'$ & 1,2,3 & 4,5 & 6 \\
\end{tabular}
 \end{center}

\textbf{Definition 3}
We call an alignment $H,T,b$ \emph{fair} if every improvement of
$H,T,b$ is fair.

The example alignment $H,T,b$ given above is not  fair  because the given alignment $H,T',b'$ is an improvement which  is not fair
for $h=\{4,5,6\}$. This is because
$b(\{4,5,6\})=\{6\}$ and $IoU(\{4,5,6\},\{6\})= \frac{1}{3}$, but in
$T'$ there now is the new segment $\{4,5\}$ and
$IoU(\{4,5,6\},\{4,5\})=\frac{2}{3}$.

\textbf{Theorem 1} 
Let $H$ and $T$ be two segmentations of the same set and $B\subseteq H\times T$. Then the following are equivalent:
\begin{itemize}
    \item all $(h,t)\in B$ satisfy (*);
    \item $B$ is a partial bijection and $H,T,B$ is a fair alignment.
\end{itemize}


\textbf{Proof}
$(\Downarrow)$ Assume that all $(h,t)\in B$ satisfy (*). By Claim~2, $B$ is a partial bijection, so we will write $B$ as the function $b$. Now suppose to the contrary that $H,T,b$ is not a fair alignment. Then 
there is an improvement of $H,T,b$ which is not fair. There are two cases. We do one, and let the improvement be $H,T',b'$ with  a $h\in H$ and a  $t\in T'$ such that $\vert h\cap t\vert \geq \vert h\cap b'(h)\vert$. Because $H,T',b'$ is an improvement, it holds that $ \vert h\cap b'(h)\vert \geq  \vert h\cap b(h)\vert $. Thus we have that $\vert h\cap t\vert \geq
\vert h\cap b(h)\vert $.

Now $t$ and $b'(h)$ are disjoint, so  $h\setminus b'(h)\supseteq h\cap t$. Because $b'$ is an improvement,   $h\cap b'(h) \supseteq h\cap b(h)$ and thus $h\setminus b'(h) \subseteq h\setminus b(h)$, 
and thus $h\setminus b(h)\supseteq h\cap t$. By chaining, we obtain $\vert h\setminus b(h)\vert \geq  \vert h \cap b(h)\vert $
which contradicts (*).

\medskip

$(\Uparrow)$
Assume $H,T,b$ is a fair alignment and  $b$ a partial bijection. 
Suppose to the contrary that (*) does not hold. Then one of the two conjuncts fails. Suppose
the first. Thus there is a $h\in H$ such that
$\vert  h\setminus b(h) \vert \geq \vert h\cap b(h) \vert$. 
Let $z=h\setminus b(h)$.
Now
create $H,T',b'$ as follows.  
\[T'=\{z\}\cup \ \{t\in T\mid t\cap z=\emptyset\}\cup \{t\setminus z\mid t\in T \mathit{\ and\ } t\cap z \neq \emptyset\},\]
and for all $h\in dom(b)$ set $b'(h)=b(h)$ if $b(h)\in T$, and
$b(h)\setminus z$ otherwise.

We will show that $H,T',b'$ is an improvement which is not fair for $h$, our required contradiction.
Because $T$ and $b$ have these properties, also  $T'$ is a partial partition, and $b'$ a partial bijection.
To show that  $H,T',b'$ is an improvement of $H,T,b$, we must show that for all $\bar{h}\in H$, the overlap with its match remained the same or increased, i.e., $\bar{h}\cap b'(\bar{h}) \supseteq \bar{h}\cap b(\bar{h})$. 
This holds by definition of $b'$ when $b(\bar{h})$ and $z$ do not overlap. Thus in particular for the segment $h$. If they do overlap, then as $z=h\setminus b(h)\subseteq h$, for all $\bar{h}\neq h$, the overlap will increase because the elements in $z$ are disjoint from $\bar{h}$ and thus taking them out of $b(\bar{h})$ reduces the number of errors.

Now we show that   $H,T',b'$ is not fair for $h$, precisely because of the set
$z \in T'$ which is not in $ran(b')$.
By definition $z=h\setminus b(h)$ and thus $z=h\cap z$. 
By assumption on $h$, $\vert  h\setminus b(h) \vert \geq \vert h\cap b(h) \vert$, and thus 
 $\vert h\cap z\vert \geq \vert h\setminus b(h)\vert$. 
We found our  desired
contradiction.



\section{Weighted P, R and F1}

The idea of weighting the F1 metric with the mean similarity between the true positives segment pairs ---which is what PQ does--- can be extended to Precision and Recall as well. In order to avoid confusion or complex naming, we   use the \emph{weighted} prefix. For completeness, we give the definitions. 

\textbf{Definition 4} Let $H$ and $T$ be two segmentations of the set $D$. Thus both $T$ and $H$ are subsets of $\mathcal{P}(D)$ in which all sets are pairwise disjoint. Let $TP\subseteq H\times T$ be a partial bijection (thus for each $h\in H$, there is at most one $(h,t)\in TP$, and similarly for $t\in T$).
Define 
\[ wTP =  {\sum }\{IoU(h,t) \mid (h,t) \in TP\}. \]
Then define, relative to this $T,H$ and $TP$, the weighted precision, recall and F1 as 

\[ P= \frac{ wTP}{\vert H\vert}, \ \ \ \ \ 
R= \frac{  wTP }{\vert T\vert},  \ \ \ \ \ 
F1= \frac{  wTP }{\vert TP \vert + .5\cdot((\vert H\vert - \vert TP \vert)  + (\vert T\vert - \vert TP \vert ) )}.
\]
\cite{kiri:pano19} call the mean IoU of the True Positives, $\frac{wTP}{\vert TP\vert}$, the \emph{segmentation quality}.
When we multiply the traditional versions of $P$, $R$ and $F1$ with $\vert TP\vert$ in the numerator with the segmentation quality, we get the weighted versions. 


\section{Empirical evaluation}

We briefly evaluate the differences between  the original $IoU>.5$, and the new $(*)$ manner of defining the aligment between predicted and true segments. We plan to do additional experiments on image data. The new definition is weaker, and thus it can lead to more TP's, but those extra TP's then have an $IoU$ value between $.33$ and $.5$, so the segmentation quality (SQ), which is the mean IoU of the TP's, will be lower.  
The recognition quality (the unweighted metrics) will go up, because we get more TP's. And the PQ (the weighted metrics) can go both ways, precisely because they form the product of the segmentation quality and the unweighted metrics. 

If there is a difference in scores,  this must come from those extra TP's. So these are segments which are hard to predict correctly, either because the method finds them hard, or because they are hard. So we expect little or no difference when systems score  high, and we expect to see differences when the task is hard (and even good systems score low on it).

Our results show actually very small differences, also on a hard task with  a state of the art segmenter based on a transformer model (BERT). On a synthetic example we obtain a bit larger differences.

In the experiments below we compare  the weighted and unweighted versions of the metrics for the two ways of creating the alignment, the original $IoU>.5$ and the new $(*)$ manner. As their difference is really in the operator we apply to the sets of missing and spurious pixels ---adding them, or having two checks---, we differentiate them by a superscript $+$ and $\&$, like in $wP^+$ and $SQ^{\&}$.

\subsection{Page stream segmentation experiment with BERT}

We do a page stream segmentation (PSS) experiment   with a segmenter based on a BERT transformer model developed in \cite{guha:mult22}.
In this task one must segment a linearly ordered stream of pages into documents. The best performing systems  learn a binary page classifier which predicts  either first or internal  page, which is enough to do  segmentation on this data. Systems work either on the page represented as an image, or as text, or using both. Here we have a system which uses the text. This is a state of the art system, which achieves mean $.77$ weighted $F1$ score  when evaluated on a dataset of 34 streams with 6.347 segments over 25K pages.  The classifier was trained on    76 streams having  17.834 segments over   63.815 pages from the same distribution. These are the findings:
\begin{itemize}
    \item of the  6.347 segments,  5.532 were TP's with $IoU>.5$, \emph{and only 4 more } with the $(*)$ matching;
    \item mean scores over these 34 streams: $SQ^{\&}=.965$ and $SQ^+=.966$, 	$F1^{\&}=.794$ and $F1^+=.793$, leading to a weighted $F1$ score of .772 for both. 
\end{itemize}
We make the task harder by testing the same classifier on a test set from another distribution than it was trained on. 
The scores drop a lot, but still hardly any difference between the two ways of alignment:
\begin{itemize}
    \item    of the 1.404 segments, 590 were TP's with the original definition and only 12 more with the $(*)$ matching;
    \item mean scores over 108 streams: $SQ^{\&}=.861$ and 	$SQ^{+}=.868$; 	$F1^{\&}=.487$ and $F1^{+}=	.479$; and  $wF1^{\&}=.433$ and 	$wF1^{+}=.429$.	
\end{itemize}
In conclusion, the scores behave as we expected, although we expected to see a larger difference in weighted $F1$ on the hard task than an insignificant $.004$ point.

\hypertarget{synthetic-experiments}{%
\subsection{Synthetic experiment}\label{synthetic-experiments}}

The next table shows the distributions of $SQ$, $F1$  and $wF1$ defined with the
two  alignments, for all 16.384
segmentations of the first 15 digits, and this true segmentation:
\[[[1,2], [3,4,5], [6,7],[8],[9], [10,11,12], [13,14],[15]].\] 
Both weighted and unweighted $F1^{\&}$
values are always higher but not very much. We can think of the table as giving the distribution of scores when predicted  with 16.384 different systems. Note that we see an increase of 229 systems finding at least one TP (1.5\% of all systems; see the counts in the columns $SQ^+$ and $SQ^{\&}$). The differences in unweighted $F1$ are 3 percent points and in weighted $F1$ just $1.5$ percent point. 

\begin{table}[t]
    \centering
    \begin{tabular}{lrrrrrr}
\toprule
{} &         $SQ^+$ &        $SQ^{\&}$ &         $F1^+$ &        $F1^{\&}$ &        $wF1^{+}$  &       $wF1^{\&}$  \\
\midrule
count &  15.556  &  15.885  &  16.384  &  16.384  &  16.384  &  16.384  \\
mean  &      0.855 &      0.819 &      0.348 &      0.379 &      0.298 &      0.314 \\
std   &      0.107 &      0.117 &      0.184 &      0.180 &      0.164 &      0.161 \\
min   &      0.600 &      0.500 &      0.000 &      0.000 &      0.000 &      0.000 \\
25\%   &      0.787 &      0.750 &      0.235 &      0.250 &      0.185 &      0.196 \\
50\%   &      0.867 &      0.833 &      0.333 &      0.375 &      0.292 &      0.302 \\
75\%   &      0.920 &      0.889 &      0.471 &      0.500 &      0.407 &      0.419 \\
max   &      1.000 &      1.000 &      1.000 &      1.000 &      1.000 &      1.000 \\
\bottomrule
\end{tabular}
    \caption{Descriptive statistics on the synthetic dataset from Section 6.2.}
    \label{tab:my_label}
\end{table}

\section{Conclusion}
We found a useful, simple, interpretable and effectively computable definition for aligning true and predicted segments which is both a necessary and sufficient condition for the alignment being a partial bijection.  
If, given a predicted and true segment,  we let TP, FP and FN stand for the pixels in the overlap, the missed and the spurious pixels, respectively, then  the necessary condition aligns the two segments if  
\textbar TP\textbar{} \textgreater{} \textbar FN\textbar{} and
\textbar TP\textbar{} \textgreater{} \textbar FP\textbar{}.
This in contrast to the $IoU>.5$ condition which translates as \textbar TP\textbar{} \textgreater{}
\textbar FN\textbar{} +\textbar FP\textbar.

Contrary to our expectations, the effect of the weaker condition was very small, at least when measured on a text segmentation task. It would be nice to see this tested on images as well. But even though the effect is small on these examples, the new condition is elegant, and more (even most)  general, so we recommend and hope it will be used in future implementations of PQ.

\bibliographystyle{unsrtnat}
\bibliography{references}  






\end{document}